\documentclass[final]{cvpr}

\usepackage{times}
\usepackage{epsfig}
\usepackage{graphicx}
\usepackage{amsmath}
\usepackage{amssymb}

\usepackage{subcaption}
\usepackage{multirow}
\usepackage{bm}
\usepackage{ifthen}
\newcommand{\redbf}[1]{{\textbf{\color{red}{#1}}}} 
\newcommand{\blueud}[1]{{\underline{\color{blue}{#1}}}} 

\usepackage[pagebackref=true,breaklinks=true,colorlinks,bookmarks=false]{hyperref}

\def\cvprPaperID{1193} 
\def\confYear{CVPR 2021}
\begin{document}

\title{\vspace{-0.6cm}Towards Real-World Blind Face Restoration with Generative Facial Prior}

\vspace{-0.4cm}
\author{
	Xintao Wang \hspace{9pt} Yu Li \hspace{9pt} Honglun Zhang \hspace{9pt} Ying Shan \\
	\vspace{-0.05cm}
	{Applied Research Center (ARC), Tencent PCG} \\
	{\tt \{xintaowang, ianyli, honlanzhang, yingsshan\}@tencent.com}\\
	\url{https://github.com/TencentARC/GFPGAN}
}

\newboolean{putfigfirst}

\setboolean{putfigfirst}{true}
\ifthenelse{\boolean{putfigfirst}}{
	
	\twocolumn[{%
		\renewcommand\twocolumn[1][]{#1}%
		\vspace{-0.5em}
		\maketitle\thispagestyle{empty}
		\begin{center}
			\centering 
			\vspace{-0.4in}
			\includegraphics[width=0.95\linewidth]{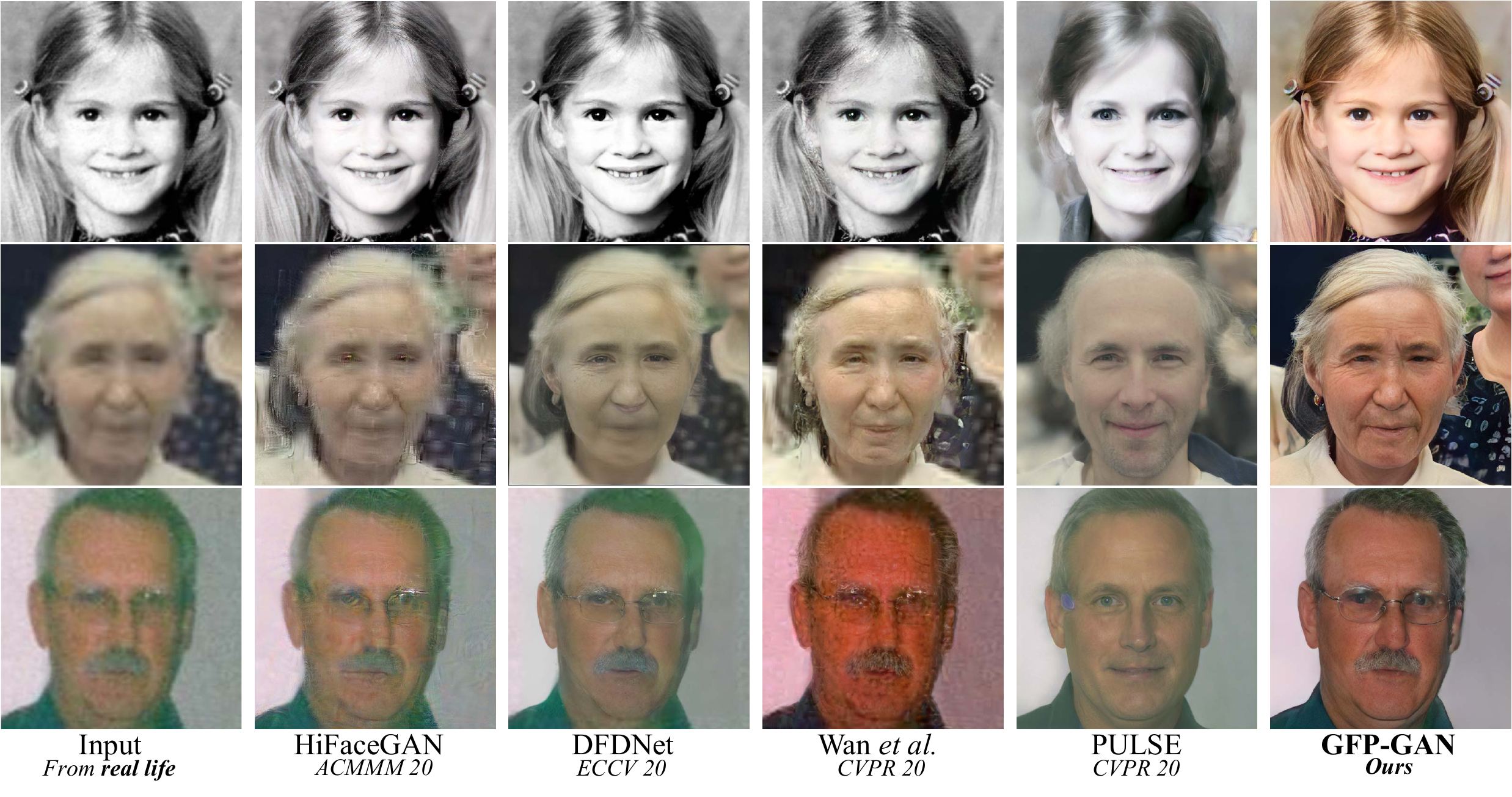}
			\vspace{-0.55cm}
			\captionof{figure}{Comparisons with state-of-the-art face restoration methods: HiFaceGAN~\cite{yang2020hifacegan}, DFDNet~\cite{li2020dfdnet}, Wan \etal~\cite{wan2020bringing} and PULSE~\cite{menon2020pulse} on the real-world low-quality images. While previous methods struggle to restore faithful facial details or retain face identity, our proposed GFP-GAN achieves a good balance of realness and fidelity with much fewer artifacts. In addition, the powerful generative facial prior allows us to  perform restoration and color enhancement jointly. (\textbf{Zoom in for best view})}
			\label{fig:teaser}
		\end{center}%
	}]
}
{
	\maketitle\thispagestyle{empty}
}

\begin{abstract}
\vspace{-1.2em}
Blind face restoration usually relies on facial priors, such as facial geometry prior or reference prior, to restore realistic and faithful details. 
However, very low-quality inputs cannot offer accurate geometric prior while high-quality references are inaccessible, limiting the applicability in real-world scenarios. 
In this work, we propose GFP-GAN that leverages rich and diverse priors encapsulated in a pretrained face GAN for blind face restoration. 
This Generative Facial Prior (GFP) is incorporated into the face restoration process via spatial feature transform layers, which allow our method to achieve a good balance of realness and fidelity. 
Thanks to the powerful generative facial prior and delicate designs, our GFP-GAN could jointly restore facial details and enhance colors with just a single forward pass, while GAN inversion methods require image-specific optimization at inference.
Extensive experiments show that our method achieves superior performance to prior art on both synthetic and real-world datasets.
%
\end{abstract}

\vspace{-2em}
\section{Introduction}
\vspace{-0.5em}
\ifthenelse{\boolean{putfigfirst}}
{}
{
\begin{figure*}[t]
	\begin{center}
		\includegraphics[width=\linewidth]{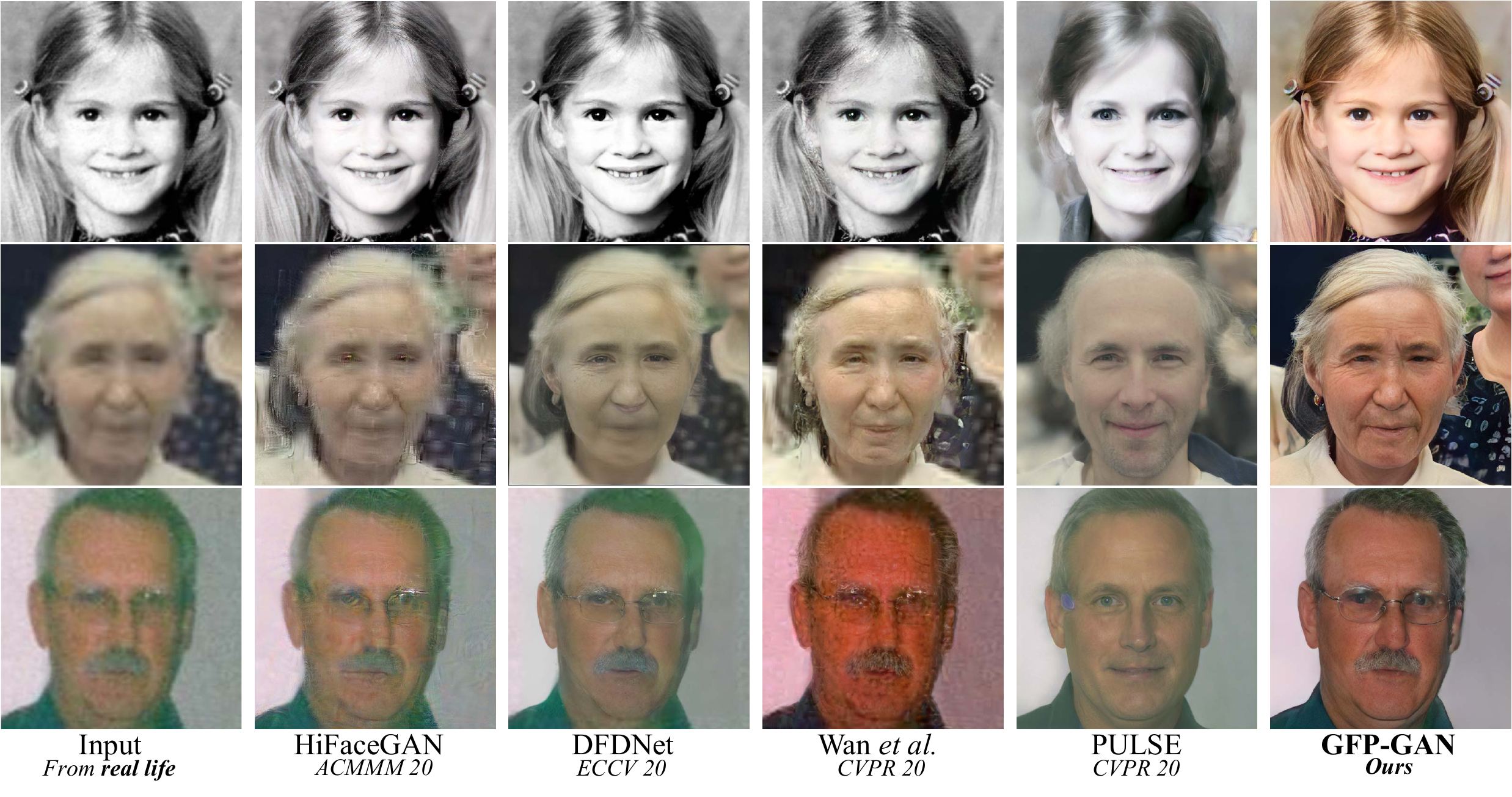}
		\caption{DNI is capable of generating continuous imagery effect transitions. (\textit{$1st$ row}) 
			from MSE effect to GAN effect in super-resolution; (\textit{$2nd$ row}) from Van Gogh style to C\'ezanne 
			style; (\textit{$3rd$ row}) from day photo to night one; (\textit{$4th$ row}) from deep depth of field 
			(DoF) to shallow one. More applications are provided in Sec.~\ref{sec:applications}. (\textbf{Zoom in for 
			best view})}
		\label{fig:teaser}
	\end{center}
	\vspace{-0.5cm}
\end{figure*}
}

Blind face restoration aims at recovering high-quality faces from the low-quality counterparts suffering from unknown degradation, such as low-resolution~\cite{dong2014learning,lim2017edsr,chen2018fsrnet}, noise~\cite{zhang2017beyond}, blur~\cite{kupyn2018deblurgan,shen2018deep}, compression artifacts~\cite{dong2015compression}, \etc. 
When applied to real-world scenarios, it becomes more challenging, due to more complicated degradation, diverse poses and expressions.
Previous works~\cite{chen2018fsrnet,yu2018face,chen2020psfrgan} typically exploit face-specific priors in face restoration, such as facial landmarks~\cite{chen2018fsrnet}, parsing maps~\cite{chen2020psfrgan,chen2018fsrnet}, facial component heatmaps~\cite{yu2018face}, and show that those \textit{geometry facial priors} are pivotal to recover accurate face shape and details. 
However, those priors are usually estimated from input images and inevitably degrades with very low-quality inputs in the real world. 
In addition, despite their semantic guidance, the above priors contain limited texture information for restoring facial details (\eg, eye pupil). 

Another category of approaches investigates \textit{reference priors}, \ie, high-quality guided faces~\cite{li2018GFRNet,li2020enhanced,dogan2019exemplar} or facial component dictionaries~\cite{li2020dfdnet}, to generate realistic results and alleviate the dependency on degraded inputs. 
However, the inaccessibility of high-resolution references limits its practical applicability, while the limited capacity of dictionaries restricts its diversity and richness of facial details. 

In this study, we leverage \textit{Generative Facial Prior} (GFP) for real-world blind face restoration, \ie, the prior implicitly encapsulated in pretrained face Generative Adversarial Network (GAN)~\cite{goodfellow2014gan} models such as StyleGAN~\cite{karras2018stylegan,karras2020stylegan2}.
These face GANs are capable of generating faithful faces with a high degree of variability, and thereby providing rich and diverse priors such as geometry, facial textures and colors, making it possible to jointly restore facial details and enhance colors (Fig.~\ref{fig:teaser}). 
%
However, it is challenging to incorporate such generative priors into the restoration process. 
Previous attempts typically use GAN inversion~\cite{gu2020mGANprior,pan2020dgp,menon2020pulse}. They first `invert' the degraded image back to a latent code of the pretrained GAN, and then conduct expensive image-specific optimization to reconstruct images.  
Despite visually realistic outputs, they usually produce images with low fidelity, as the low-dimension latent codes are insufficient to guide accurate restoration.

To address these challenges, we propose the GFP-GAN with delicate designs to achieve a good balance of realness and fidelity in a single forward pass. 
Specifically, GFP-GAN consists of a degradation removal module and a pretrained face GAN as facial prior. They are connected by a direct latent code mapping, and several Channel-Split Spatial Feature Transform (CS-SFT) layers in a coarse-to-fine manner. 
The proposed CS-SFT layers perform \textit{spatial} modulation on a split of features and leave the left features to directly pass through for better information preservation, allowing our method to effectively incorporate generative prior while retraining high fidelity.
Besides, we introduce facial component loss with local discriminators to further enhance perceptual facial details, while employing identity preserving loss to further improve fidelity. 

We summarize the contributions as follows.
(1) We leverage rich and diverse generative facial priors for blind face restoration. Those priors contain sufficient facial textures and color information, allowing us to jointly perform face restoration and color enhancement. 
(2) We propose the GFP-GAN framework with delicate designs of architectures and losses to incorporate generative facial prior. Our GFP-GAN with CS-SFT layers achieves a good balance of fidelity and texture faithfulness in a single forward pass.  
(3) Extensive experiments show that our method achieves superior performance to prior art on both synthetic and real-world datasets.

\vspace{-0.5em}
\section{Related Work}
\vspace{-0.5em}
\noindent\textbf{Image Restoration} typically includes super-resolution~\cite{dong2014learning,lim2017edsr,timofte2017ntire,liu2018non,zhang2018rcan,yu2019path,guo2020closed,liu2020residual}, denoising~\cite{zhang2017beyond,lefkimmiatis2017non,helou2020stochastic}, deblurring~\cite{xu2014deep,kupyn2018deblurgan,shen2018deep} and  compression removal~\cite{dong2015compression,guo2016building}.
To achieve visually-pleasing results, generative adversarial network~\cite{goodfellow2014gan} is usually employed as loss supervisions to push the solutions closer to the natural manifold~\cite{ledig2017srgan,sajjadi2017enhancenet,wang2018esrgan,chen2018image,galteri2017deep}, while our work attempts to leverage the pretrained face GANs as generative facial priors (GFP).
%

\noindent\textbf{Face Restoration.}
Based on general face hallucination~\cite{cao2017attention,huang2017wavelet,xu2017learning,yu2018super}, two typical face-specific priors: geometry priors and reference priors, are incorporated to further improve the performance. 
The geometry priors include facial landmarks~\cite{chen2018fsrnet,kim2019progressive,zhu2016deep}, face parsing maps~\cite{shen2018deep,chen2020psfrgan,chen2018fsrnet} and facial component heatmaps~\cite{yu2018face}.
However, 1) those priors require estimations from low-quality inputs and inevitably degrades in real-world scenarios. 2) They mainly focus on geometry constraints and may not contain adequate details for restoration. Instead, our employed GFP does not involve an explicit geometry estimation from degraded images, and contains adequate textures inside its pretrained network.
 
Reference priors~\cite{li2018GFRNet,li2020enhanced,dogan2019exemplar} usually rely on reference images of the same identity. 
To overcome this issue, DFDNet~\cite{li2020dfdnet} suggests to construct a face dictionary of each component (\eg, eyes, mouth) with CNN features to guide the restoration. 
However, DFDNet mainly focuses on components in the dictionary and thus degrades in the regions beyond its dictionary scope (\eg, hair, ears and face contour), instead, our GFP-GAN could treat faces as a whole to restore. Moreover, the limited size of dictionary restricts its diversity and richness, while the GFP could provide rich and diverse priors including geometry, textures and colors. 

\noindent\textbf{Generative Priors} of pretrained GANs~\cite{karras2018pggan,karras2018stylegan,karras2020stylegan2,brock2018large} is previously exploited by GAN inversion~\cite{abdal2019image2stylegan,zhu2020domain,pan2020dgp,gu2020mGANprior}, whose primary aim is to find the closest latent codes 
given an input image.
PULSE~\cite{menon2020pulse} iteratively optimizes the latent code of StyleGAN~\cite{karras2018stylegan} until the distance between outputs and inputs is below a threshold. mGANprior~\cite{gu2020mGANprior} attempts to optimize multiple codes to improve the reconstruction quality. 
However, these methods usually produce images with low fidelity, as the low-dimension latent codes are insufficient to guide the restoration. In contrast, our proposed CS-SFT modulation layers enable prior incorporation on multi-resolution spatial features to achieve high fidelity. 
Besides, expensive iterative optimization is not required in our GFP-GAN during inference.

\begin{figure*}
	\vspace{-0.6cm}
	\begin{center}
		\includegraphics[width=\linewidth]{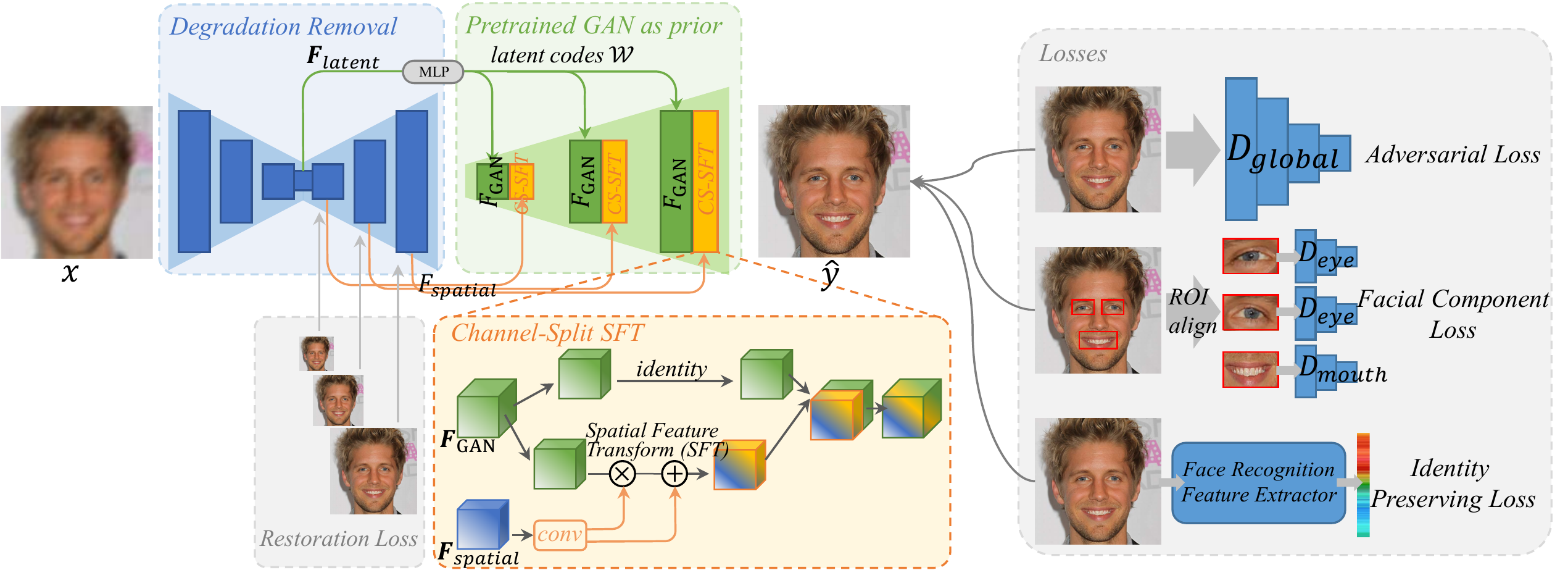}
	\end{center}
	\vspace{-0.5cm}
	\caption{\textbf{Overview of GFP-GAN framework}. It consists of a degradation removal module (U-Net) and a pretrained face GAN as facial prior. They are bridged by a latent code mapping and several Channel-Split Spatial Feature Transform (CS-SFT) layers. During training, we employ 1) intermediate restoration losses to remove complex degradation, 2) Facial component loss with discriminators to enhance facial details, and 3) identity preserving loss to retain face identity.}
	\label{fig:overview}
	\vspace{-0.5cm}
\end{figure*}

\noindent\textbf{Channel Split Operation} is usually explored to design compact models and improve model representation ability.
MobileNet~\cite{howard2017mobilenets} proposes depthwise convolutions and 
GhostNet~\cite{han2020ghostnet} splits the convolutional layer into two parts and uses
fewer filters to generate intrinsic feature maps. 
Dual path architecture in DPN~\cite{chen2017dpn} enables feature re-usage and new feature exploration for each path, thus improving its representation ability. A similar idea is also employed in super-resolution~\cite{zhao2019channel}.
Our CS-SFT layers share the similar spirits, but with different operations and purposes.
We adopt spatial feature transform~\cite{wang2018sftgan,park2019spade} on one split and leave the left split as identity to achieve a good balance of realness and fidelity.

\noindent\textbf{Local Component Discriminators.}
Local discriminator is proposed to focus on local patch distributions~\cite{iizuka2017globally,li2017generative,wang2017pix2pixHD}.
%
When applied to faces, those discriminative losses are imposed on separate semantic facial regions~\cite{li2018beautygan,gu2019ladn}. 
Our introduced facial component loss also adopts such designs but with a further style supervision based on the learned discriminative features.


\section{Methodology}

\subsection{Overview of GFP-GAN}
We describe GFP-GAN framework in this section. Given an input facial image $\bm{x}$ suffering from unknown degradation, the aim of blind face restoration is to estimate a high-quality image $\hat{\bm{y}}$, which is as similar as possible to the ground-truth image $\bm{y}$, in terms of realness and fidelity.

The overall framework of GFP-GAN is depicted in Fig.~\ref{fig:overview}. GFP-GAN is comprised of a degradation removal module (U-Net) and a pretrained face GAN (such as StyleGAN2~\cite{karras2020stylegan2}) as prior. They are bridged by a latent code mapping and several Channel-Split Spatial Feature Transform (CS-SFT) layers. 
Specifically, the degradation removal module is designed to remove complicated degradation, and extract two kinds of features, \ie 1) latent features $\bm{F}_{latent}$ to map the input image to the closest latent code in StyleGAN2, and 2) multi-resolution spatial features $\bm{F}_{spatial}$ for modulating the StyleGAN2 features. 

After that, $\bm{F}_{latent}$ is mapped to intermediate latent codes $\mathcal{W}$ by several linear layers. Given the close latent code to the input image, StyleGAN2 could generate \textit{intermediate convolutional features}, denoted by $\bm{F}_{\text{GAN}}$. These features provide rich facial details captured in the weights of pre-trained GAN.
Multi-resolution features $\bm{F}_{spatial}$ are used to spatially modulate the face GAN features $\bm{F}_{\text{GAN}}$ with the proposed CS-SFT layers in a coarse-to-fine manner, achieving realistic results while preserving high fidelity.

During training, except for the global discriminative loss, we introduce facial component loss with discriminators to enhance the perceptually significant face components, \ie, eyes and mouth. In order to retrain identity, we also employ identity preserving guidance.

\subsection{Degradation Removal Module}
\label{method:degradation_removal}
Real-world blind face restoration faces with complicated and severer degradation, which is typically a mixture of low-resolution, blur, noise and JPEG artifacts. 
The degradation removal module is designed to explicitly remove the above degradation and extract `clean' features $\bm{F}_{latent}$ and $\bm{F}_{spatial}$, alleviating the burden of subsequent modules.
We adopt the U-Net~\cite{ronneberger2015unet} structure as our degradation remove module, as it could 1) increase receptive field for large blur elimination, and 2) generate multi-resolution features.
The formulation is as follows:
\begin{equation}\label{equ:unet}
	\bm{F}_{latent}, \bm{F}_{spatial} = \mathtt{U\text{-}Net}(\bm{x}).
\end{equation}
The latent features $\bm{F}_{latent}$ is used to map the input image to the closest latent code in StyleGAN2 (Sec.~\ref{method:latent_code_mapping}). The multi-resolution spatial features $\bm{F}_{spatial}$ are used to modulate the StyleGAN2 features (Sec.~\ref{method:cs-sft}). 

In order to have an intermediate supervision for removing degradation, we employ the L1 restoration loss in each resolution scale in the early stage of training. Specifically, we also output images for each resolution scale of the U-Net decoder, and then restrict these outputs to be close to the pyramid of the ground-truth image.
%

%
%


\subsection{Generative Facial Prior and Latent Code \\ Mapping}
\label{method:latent_code_mapping}
A pre-trained face GAN captures a distribution over faces in its leaned weights of convolutions, namely, generative prior~\cite{gu2020mGANprior,pan2020dgp}. 
We leverage such pretrained face GANs to provide diverse and rich facial details for our task.
A typical way of deploying generative priors is to map the input image to its closest latent codes $Z$, and then generate  the corresponding output by a pretrained GAN~\cite{abdal2019image2stylegan,zhu2020domain,pan2020dgp,gu2020mGANprior}.
However, these methods usually require time-consuming iterative optimization for preserving fidelity.
Instead of producing a final image directly, we generate \textit{intermediate convolutional features} $\bm{F}_{\text{GAN}}$ of the closest face, as it contains more details and could be further modulated by input features for better fidelity (see Sec.~\ref{method:cs-sft}).

Specifically, given the encoded vector $\bm{F}_{latent}$ of the input image (produced by the U-Net, Eq.~\ref{equ:unet}), we first map it to intermediate latent codes $\mathcal{W}$  for better preserving semantic property~ \ie, the intermediate space transformed from $Z$ with several multi-layer perceptron layers (MLP)~\cite{zhu2020domain}.
The latent codes $\mathcal{W}$ then pass through each convolution layer in the pre-trained GAN, and generate GAN features for \textit{each resolution scale}. 
\begin{equation}
	\label{equ:gfp}
	\begin{split}
		\mathcal{W} &= \mathtt{MLP}(\bm{F}_{latent}),\\
		\bm{F}_{\text{GAN}} &= \mathtt{StyleGAN}(\mathcal{W}).
	\end{split}
\end{equation}

%
%

%
%

\noindent\textbf{Discussion: Joint Restoration and Color Enhancement}.
Generative models capture diverse and rich priors beyond realistic details and vivid textures. 
For instance, they also encapsulate \textit{color} priors, which could be employed in our task for joint face restoration and color enhancement.
Real-world face images, \eg, old photos, usually have black-and-white color, vintage yellow color, or dim color. Lively color prior in generative facial prior allows us to perform color enhancement including colorization~\cite{zhang2016colorful}.
We believe the generative facial priors also incorporate conventional geometric priors~\cite{chen2018fsrnet,yu2018face}, 3D priors~\cite{gecer2019ganfit}, \etc for restoration and manipulation.

\subsection{Channel-Split Spatial Feature Transform}
\label{method:cs-sft}
In order to better preserve fidelity, we further use the input spatial features $\bm{F}_{spatial}$ (produced by the U-Net, Eq.~\ref{equ:unet}) to modulate the 
GAN features $\bm{F}_{\text{GAN}}$ from Eq.~\ref{equ:gfp}.
Preserving spatial information from inputs is crucial for face restoration, as it usually requires local characteristics for fidelity preservation, and adaptive restoration at different spatial locations of a face.
Therefore, we employ Spatial Feature Transform (SFT)~\cite{wang2018sftgan}, which generates affine transformation parameters for spatial-wise feature modulation, and has shown its effectiveness in incorporating other conditions in image restoration~\cite{wang2018sftgan,li2020dfdnet} and image generation~\cite{park2019spade}.

Specifically, at each resolution scale, we generate a pair of affine transformation parameters $(\bm{\alpha}, \bm{\beta})$ from input features $\bm{F}_{spatial}$ by several convolutional layers. After that, the modulation is carried out by scaling and shifting the GAN features $\bm{F}_{\text{GAN}}$, formulated by:
\begin{equation}
	\vspace{-0.5em}
	\label{equ:sft}
	\begin{split}
		\bm{\alpha}, \bm{\beta} &= \mathtt{Conv}(\bm{F}_{spatial}),\\
		\bm{F}_{output}& = \mathtt{SFT}(\bm{F}_{\text{GAN}} | \bm{\alpha}, \bm{\beta}) = \bm{\alpha} \odot \bm{F}_{\text{GAN}} +  \bm{\beta}.
	\end{split}
\end{equation}

%
To achieve a better balance of realness and fidelity, we further propose channel-split spatial feature transform (CS-SFT) layers, which perform spatial modulation on part of the GAN features by input features $\bm{F}_{spatial}$ (contributing to fidelity) and leave the left GAN features (contributing to realness) to directly pass through, as shown in Fig.~\ref{fig:overview}:
%
\vspace{-0.8em}
\begin{align}\vspace{-0.5em}\label{equ:cs-sft}
	\bm{F}_{output} &=\mathtt{CS}\text{-}\mathtt{SFT}(\bm{F}_{\text{GAN}} | \bm{\alpha}, \bm{\beta}) \\
	&=\mathtt{Concat} [\mathtt{Identity}(\bm{F}_{\text{GAN}}^{split0}), \bm{\alpha} \odot \bm{F}_{\text{GAN}}^{split1} +  \bm{\beta}], \nonumber
\end{align} 
where $\bm{F}_{\text{GAN}}^{split0}$ and $\bm{F}_{\text{GAN}}^{split1}$ are split features from $\bm{F}_{\text{GAN}}$ in channel dimension, and $\mathtt{Concat}[\cdot,\cdot]$ denotes the concatenation operation.

As a result, CS-SFT enjoys the benefits of directly incorporating prior information and effective modulating by input images, thereby achieving a good balance between texture faithfulness and fidelity.
%
Besides, CS-SFT could also reduce complexity as it requires fewer channels for modulation, similar to GhostNet~\cite{han2020ghostnet}.

We conduct channel-split SFT layers at each resolution scale, and finally generate a restored face $\hat{\bm{y}}$.

\subsection{Model Objectives}
\label{method:model_objectives}
The learning objective of training our GFP-GAN consists of: 1) reconstruction loss that constraints the outputs $\hat{\bm{y}}$ close to the ground-truth $\bm{y}$, 2) adversarial loss for restoring realistic textures, 3) proposed facial component loss to further enhance facial details, and 4) identity preserving loss. 

\noindent\textbf{Reconstruction Loss.}
We adopt the widely-used L1 loss and perceptual loss~\cite{johnson2016perceptual, ledig2017srgan} as our reconstruction loss $\mathcal{L}_{rec}$, defined as follows:
\vspace{-0.3em}
\begin{equation}\label{equ:reconstruction}
	\mathcal{L}_{rec} = \lambda_{l1} \|\hat{\bm{y}}-\bm{y} \|_1 + \lambda_{per} \|\phi(\hat{\bm{y}})-\phi(\bm{y}) \|_1,
\end{equation}
\vspace{-0.3em}
where $\phi$ is the pretrained VGG-19 network~\cite{simonyan2015vgg} and we use the $\{\mathtt{conv1,} \cdots \mathtt{,conv5}\}$ feature maps before activation~\cite{wang2018esrgan}. $\lambda_{l1}$ and $\lambda_{per}$ denote the loss weights of L1 and perceptual loss, respectively.

\noindent\textbf{Adversarial Loss.}
We employ adversarial loss $L_{adv}$ to encourage the GFP-GAN to favor the solutions in the natural image manifold and generate realistic textures. Similar to StyleGAN2~\cite{karras2020stylegan2}, logistic loss~\cite{goodfellow2014gan} is adopted:
\vspace{-0.8em}
\begin{equation}\label{equ:adversarial}
	\mathcal{L}_{adv} = -\lambda_{adv}\mathbb{E}_{\hat{\bm{y}}}\ \mathtt{softplus}(D(\hat{\bm{y}})) 
\end{equation}
where $D$ denotes the discriminator and $\lambda_{adv}$ represents the adversarial loss weight.

\noindent\textbf{Facial Component Loss.}
In order to further enhance the perceptually significant face components, we introduce facial component loss with local discriminators for left eye, right eye and mouth. 
As shown in Fig.~\ref{fig:overview}, we first crop interested regions with ROI align~\cite{he2017maskrcnn}. For each region, we train separate and small local discriminators to distinguish whether the restore patches are real, pushing the patches close to the natural facial component distributions.

Inspired by~\cite{wang2017pix2pixHD}, we further incorporate a feature style loss based on the learned discriminators. Different from previous feature matching loss with spatial-wise constraints~\cite{wang2017pix2pixHD}, our feature style loss attempts to match the Gram matrix statistics~\cite{gatys2016style} of real and restored patches. Gram matrix calculates the feature correlations and usually effectively captures texture information~\cite{gondal2018unreasonable}. 
We extract features from multiple layers of the learned local discriminators and learn to match these Gram statistic of intermediate representations from the real and restored patches.
Empirically, we found the feature style loss performs better than previous feature matching loss in terms of generating realistic facial details and reducing unpleasant artifacts.

The facial component loss is defined as follows. The first term is the discriminative loss~\cite{goodfellow2014gan} and the second term is the feature style loss:
\begin{equation}\label{equ:face_component}
\begin{split}
	\mathcal{L}_{comp} = \sum\limits_{\mathtt{ROI}}{} \lambda_{local}\ \mathbb{E}_{\hat{\bm{y}}_{\mathtt{ROI}}} [\log(1-D_{\mathtt{ROI}}(\hat{\bm{y}}_{\mathtt{ROI}}))]  + \\
	\lambda_{fs}\ \|\mathtt{Gram}(\psi(\hat{\bm{y}}_{\mathtt{ROI}})) - \mathtt{Gram}(\psi(\bm{y}_{\mathtt{ROI}})) \|_1
\end{split}
\end{equation}
where $\mathtt{ROI}$ is region of interest from the component collection $\{\mathtt{left\_eye, right\_eye, mouth}\}$.
$D_{\mathtt{ROI}}$ is the local discriminator for each region. $\psi$ denotes the multi-resolution features from the learned discriminators. $\lambda_{local}$ and $\lambda_{fs}$ represent the loss weights of local discriminative loss and feature style loss, respectively.

\noindent\textbf{Identity Preserving Loss.}
We draw inspiration from~\cite{huang2017tpgan} and apply identity preserving loss in our model.
Similar to perceptual loss~\cite{johnson2016perceptual}, we define the loss based on the feature embedding of an input face. Specifically, we adopt the pretrained  face recognition ArcFace~\cite{deng2019arcface} model, which captures the most prominent features for identity discrimination. The identity preserving loss enforces the restored result to have a small distance with the ground truth in the compact deep feature space:
\begin{equation}\label{equ:identity}
	\mathcal{L}_{id} = \lambda_{id} \|\eta(\hat{\bm{y}})-\eta(\bm{y}) \|_1,
\end{equation}
where $\eta$ represents face feature extractor, \ie ArcFace~\cite{deng2019arcface} in our implementation. $\lambda_{id}$ denotes the weight of identity preserving loss.

\medskip
The overall model objective is a combination of the above losses:
\begin{equation}\label{equ:total_loss}
	\mathcal{L}_{total} = \mathcal{L}_{rec} + \mathcal{L}_{adv} + \mathcal{L}_{comp} + \mathcal{L}_{id}.
\end{equation}
The loss hyper-parameters are set as follows: $\lambda_{l1}=0.1$, $\lambda_{per}=1$, $\lambda_{adv}=0.1$,  $\lambda_{local}=1$, $\lambda_{fs}=200$ and $\lambda_{id}=10$. 

\section{Experiments} \label{sec:experiments}

\subsection{Datasets and Implementation}
\noindent\textbf{Training Datasets.}
We train our GFP-GAN on the FFHQ dataset~\cite{karras2018stylegan}, which consists of $70,000$ high-quality images. We resize all the images to $512^2$ during training. 

Our GFP-GAN is trained on synthetic data that approximate to the real low-quality images and generalize to real-world images during inference. 
We follow the practice in~\cite{li2018GFRNet,li2020dfdnet} and adopt the following degradation model to synthesize training data:

\begin{equation}\label{equ:degradation} 
	\bm{x} = [(\bm{y}\circledast \bm{k}_{\sigma})\downarrow_{r} + \bm{n}_{\delta}]_{\mathtt{JPEG}_{q}}.
\end{equation}
The high quality image $\bm{y}$ is first convolved with Gaussian blur kernel $\bm{k}_{\sigma}$ followed by a downsampling operation with a scale factor $r$. After that, additive white Gaussian noise $\bm{n}_{\delta}$ is added to the image and finally it is compressed by JPEG with quality factor $q$.
Similar to~\cite{li2020dfdnet}, for each training pair, we randomly sample $\sigma$, $r$, $\delta$ and $q$ from $\{0.2:10\}$, $\{1:8\}$, $\{0:15\}$, $\{60:100\}$, respectively. We also add color jittering during training for color enhancement.

\begin{figure*}
	\vspace{-0.8cm}
	\begin{center}
		\includegraphics[width=\linewidth]{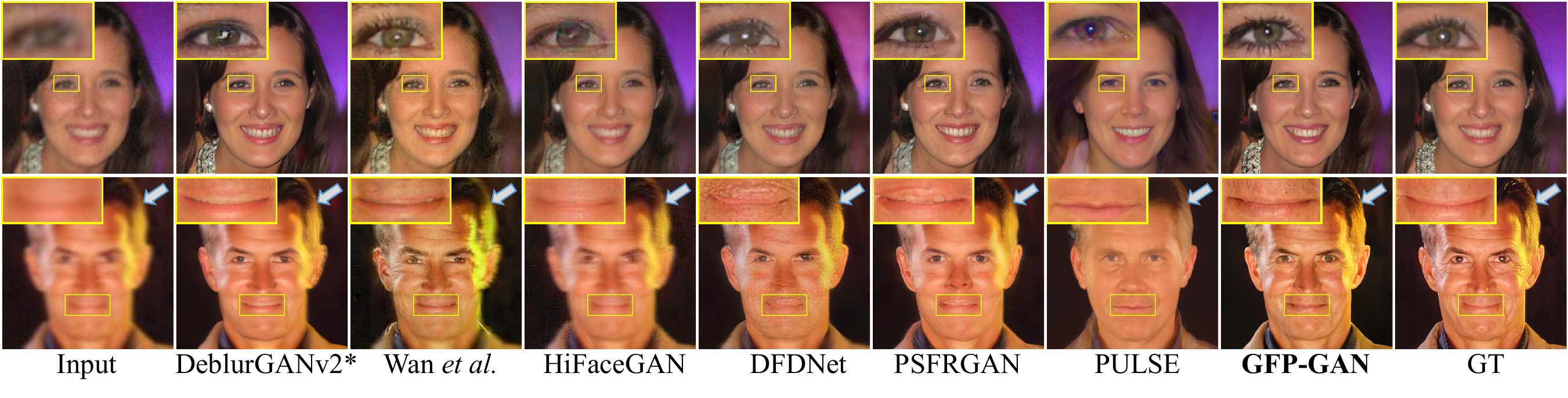}
	\end{center}
	\vspace{-0.8cm}
	\caption{Qualitative comparison on the \textbf{CelebA-Test} for blind face restoration. Our GFP-GAN produces faithful details in eyes, mouth and hair. \textbf{Zoom in for best view.}}
	\label{fig:celeb_compare}
	\vspace{-0.1cm}
\end{figure*}

\noindent\textbf{Testing Datasets.}
We construct one synthetic dataset and three different \textbf{real} datasets with distinct sources. All these datasets have no overlap with our training dataset.
We provide a brief introduction here. 

$\bullet$ \textit{CelebA-Test} is the synthetic dataset with 3,000 CelebA-HQ images from its testing partition~\cite{liu2015faceattributes}. The generation way is the same as that during training.

$\bullet$ \textit{LFW-Test.} LFW~\cite{LFWTech} contains low-quality images \textbf{in the wild}. We group all the first image for each identity in the validation partition, forming 1711 testing images.

$\bullet$ \textit{CelebChild-Test} contains 180 child faces of celebrities collected from \textbf{the Internet}. They are low-quality and many of them are black-and-white old photos.

$\bullet$ \textit{WebPhoto-Test.} We crawled 188 low-quality photos \textbf{in real life from the Internet} and extracted 407 faces to construct the WebPhoto testing dataset. These photos have diverse and complicated degradation. Some of them are old photos with very severe degradation on both details and color.

\begin{figure}[]
	\begin{center}
		\includegraphics[width=\linewidth]{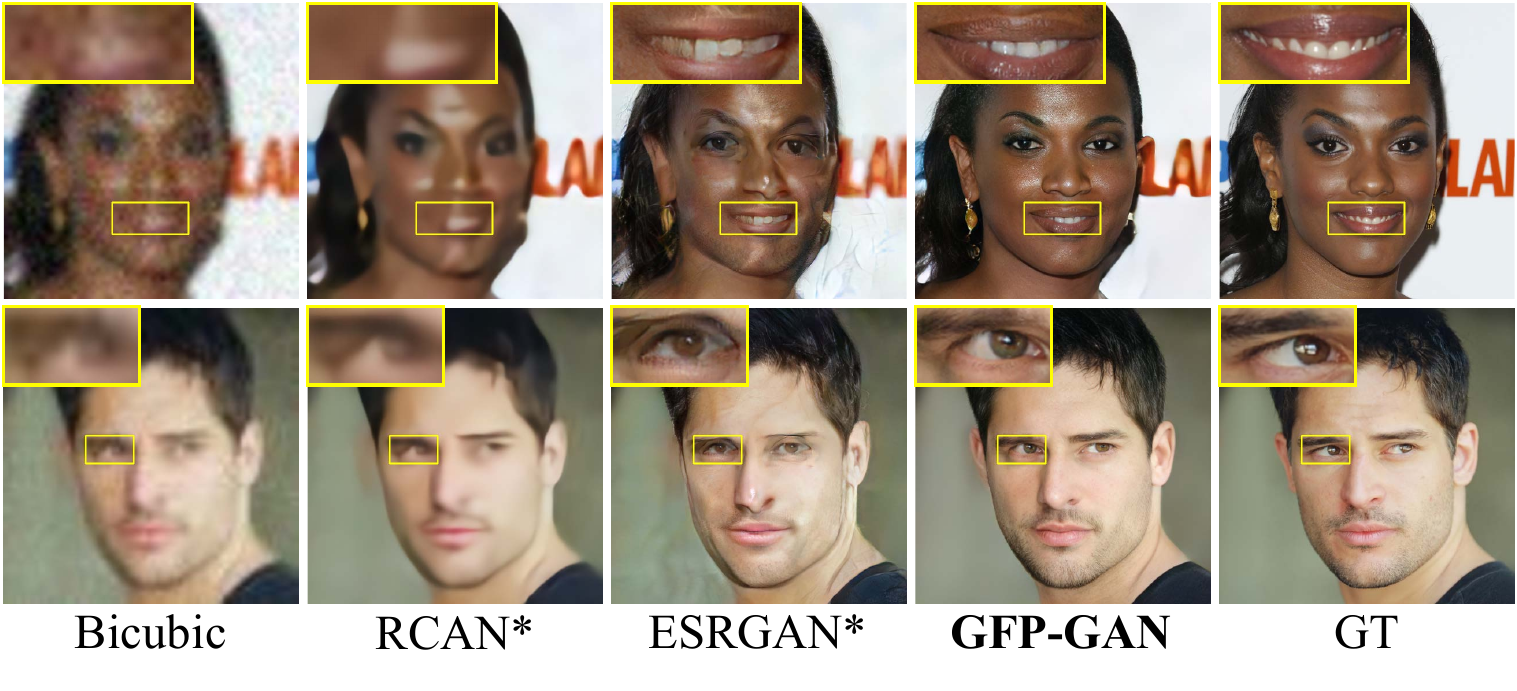}
	\end{center}
	\vspace{-0.8cm}
	\caption{Comparison on the \textbf{CelebA-Test} for $\times 4$ face super-resolution. Our GFP-GAN restores realistic teeth and faithful eye gaze direction. \textbf{Zoom in for best view.}}
	\label{fig:celeb_sr_compare}
	\vspace{-0.5cm}
\end{figure}

\noindent\textbf{Implementation.}
We adopt the pretrained StyleGAN2~\cite{karras2020stylegan2} with $512^2$ outputs as our generative facial prior.
The channel multiplier of StyleGAN2 is set to one for compact model size. 
The UNet for degradation removal consists of seven downsamples and seven upsamples, each with a residual block~\cite{he2016resnet}. For each CS-SFT layer, we use two convolutional layers to generate the affine parameters $\alpha$ and $\beta$ respectively.

%
The training mini-batch size is set to 12. We augment the training data with horizontal flip and color jittering.
We consider three components: left\_eye, right\_eye, mouth for face component loss as they are perceptually significant. 
Each component is cropped by ROI align~\cite{he2017maskrcnn} with face landmarks provided in the origin training dataset.
We train our model with Adam optimizer~\cite{kingma2014adam} for a total of $800$k iterations.
The learning rate was set to $2\times10^{-3}$ and then decayed by a
factor of 2 at the $700$k-th, $750$k-th iterations.
We implement our models with the PyTorch framework and train them using four
NVIDIA Tesla P40 GPUs.

\subsection{Comparisons with State-of-the-art Methods}
We compare our GFP-GAN with several state-of-the-art face restoration methods:  HiFaceGAN~\cite{yang2020hifacegan}, DFDNet~\cite{li2020dfdnet}, PSFRGAN~\cite{chen2020psfrgan}, Super-FAN~\cite{bulat2018super} and Wan \etal~\cite{wan2020bringing}. 
GAN inversion methods for face restoration: PULSE~\cite{menon2020pulse} and mGANprior~\cite{gu2020mGANprior} are also included for comparison. 
We also compare our GFP-GAN with image restoration methods: RCAN~\cite{zhang2018rcan}, ESRGAN~\cite{wang2018esrgan} and DeblurGANv2~\cite{kupyn2019deblurganv2}, and we finetune them on our face training set for fair comparisons.
We adopt their \textit{official} codes except for Super-FAN, for which we use a re-implementation.

\begin{table}[h]
	\vspace{-0.5cm}
	\small
	\centering
	\caption{Quantitative comparison on \textbf{CelebA-Test} for blind face restoration. \redbf{Red} and \blueud{blue} indicates the best and the second best performance. `*' denotes finetuning on our training set. Deg. represents the identity distance.}
	\label{tab:celeba_blind}
	\tabcolsep=0.1cm
	\scalebox{0.95}{
		\hspace{-0.5cm}
		\begin{tabular}{c|ccc|c|cc}
			\hline
			Methods       & LPIPS$\downarrow$  & FID$\downarrow$   &NIQE $\downarrow$    & Deg.$\downarrow$   & PSNR$\uparrow$  & SSIM$\uparrow$  \\ \hline
			Input        & 0.4866 & 143.98  & 13.440& 47.94 & 25.35  & 0.6848  \\
			DeblurGANv2*~\cite{kupyn2019deblurganv2}   &   \blueud{0.4001}    &   52.69      &   4.917      &   \blueud{39.64}     &    \redbf{25.91}    &     \redbf{0.6952}  \\
			Wan \etal.~\cite{wan2020bringing}  & 0.4826 & 67.58 & 5.356   & 43.00 & 24.71  & 0.6320  \\
			HiFaceGAN~\cite{yang2020hifacegan}      & 0.4770  & 66.09 & 4.916  & 42.18& 24.92 & 0.6195  \\
			DFDNet~\cite{li2020dfdnet}         & 0.4341 & 59.08 & \blueud{4.341}   & 40.31& 23.68  & 0.6622  \\
			PSFRGAN~\cite{chen2020psfrgan}       & 0.4240 & \blueud{47.59}  & 5.123  & 39.69 & 24.71  & 0.6557  \\ \hline
			mGANprior~\cite{gu2020mGANprior}&   0.4584    &   82.27      &   6.422      &   55.45      &    24.30     &     0.6758   \\
			PULSE~\cite{menon2020pulse}&   0.4851     &   67.56      &   5.305      &   69.55      &    21.61    &     0.6200  \\\hline
			\textbf{GFP-GAN (ours)} & \redbf{0.3646}  & \redbf{42.62}  & \redbf{4.077} & \redbf{34.60} & 25.08   & 0.6777  \\ \hline
			GT                 & 0      & 43.43 & 4.292    & 0    & $\infty$    & 1  \\ \hline
	\end{tabular}}
	\vspace{-0.5cm}
\end{table}

For the evaluation, we employ the widely-used non-reference perceptual metrics: FID~\cite{heusel2017gans} and NIQE~\cite{mittal2012making}.
We also adopt pixel-wise metrics (PSNR and SSIM) and the perceptual metric (LPIPS~\cite{zhang2018perceptual}) for the CelebA-Test with Ground-Truth (GT). 
We measure the identity distance with angels in the ArcFace~\cite{deng2019arcface} feature embedding, where smaller values indicate closer  identity to the GT.

\begin{figure*}[!t]
	\vspace{-0.5cm}
	\begin{center}
		\includegraphics[width=\linewidth]{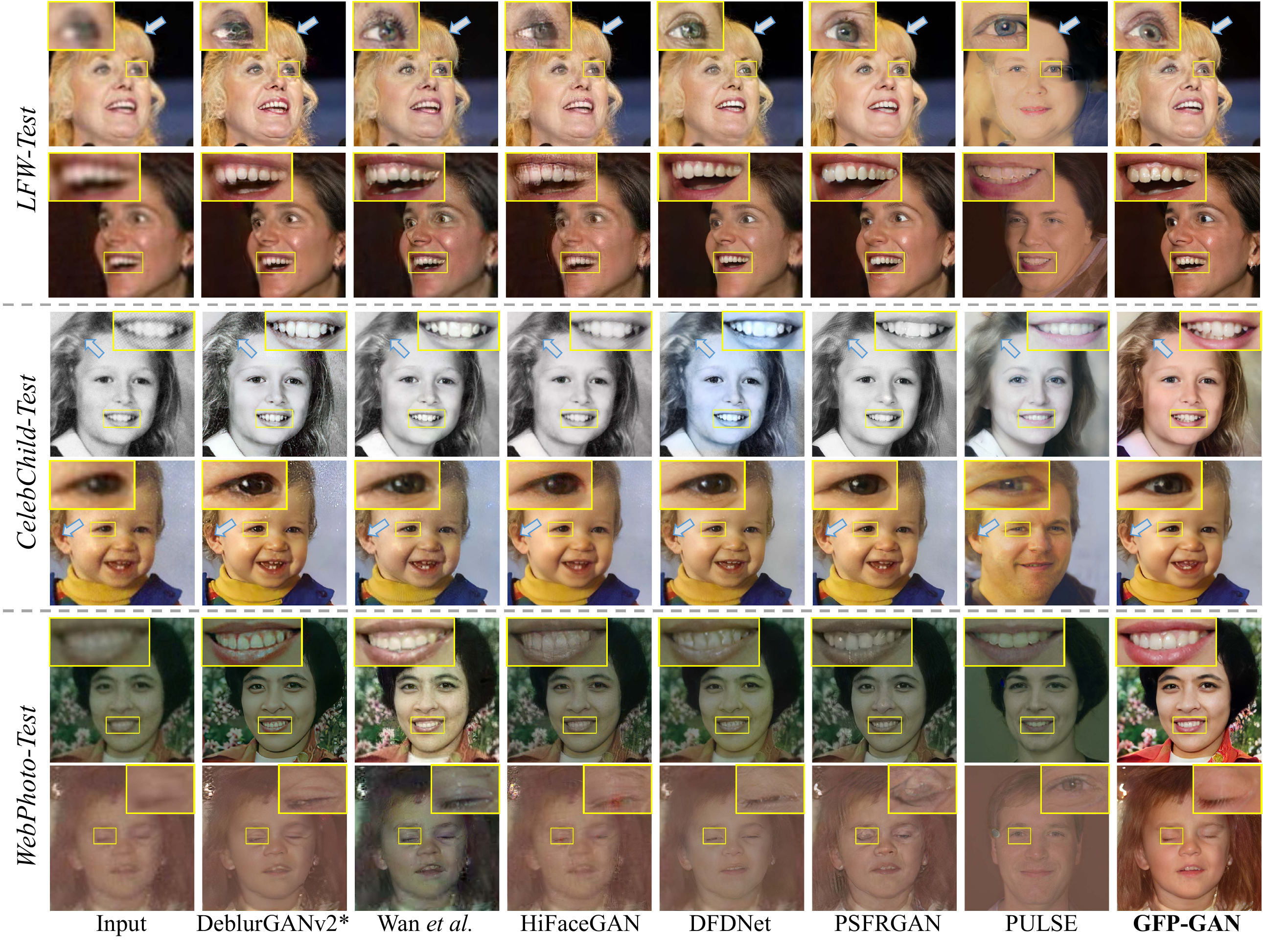}
	\end{center}
	\vspace{-0.8cm}
	\caption{Qualitative comparisons on three \textbf{real-world} datasets. \textbf{Zoom in for best view.}}
	\label{fig:real_comparison}
	\vspace{-0.5cm}
\end{figure*}


\noindent\textbf{Synthetic CelebA-Test.}
The comparisons are conducted under two settings:  1) blind face restoration whose inputs and outputs have the same resolution. 2) $4\times$ face super-resolution. Note that our method could take upsampled images as inputs for face super-resolution.

The quantitative results for each setting are shown in Table.~\ref{tab:celeba_blind} and Table.~\ref{tab:celeba_sr}. 
On both settings, GFP-GAN achieves the lowest LPIPS, indicating that our results is perceptually close to the ground-truth. GFP-GAN also obtain the lowest FID and NIQE, showing that the outputs have a close distance to the real face distribution and natural image distribution, respectively. 
Besides the perceptual performance, our method also retains better identity, indicated by the smallest degree in the face feature embedding. Note that 1) the lower FID and NIQE of our method than GT does not indicate that our performance is better than GT, as those `perceptual' metrics are well correlated with the human-opinion-scores on a coarse scale, but \textit{not always well correlated on a finer scale}~\cite{blau20182018}; 2) the pixel-wise metrics PSNR and SSIM are not correlation well with the subjective evaluation of human observers~\cite{blau20182018,ledig2017srgan} and our model is not good at these two metrics.

\begin{table}[th]
	\small
	\centering
	\caption{Quantitative comparison on \textbf{CelebA-Test} for $4\times$ face super-resolution. \redbf{Red} and \blueud{blue} indicates the best and the second best performance.  `*' denotes finetuning on our training set. Deg. represents the identity distance.}
	\vspace{-0.3cm}
	\label{tab:celeba_sr}
	\tabcolsep=0.1cm
	\scalebox{0.95}{
		\hspace{-0.5cm}
		\begin{tabular}{c|ccc|c|cc}
			\hline
			Methods        & LPIPS$\downarrow$ & FID$\downarrow$  &NIQE $\downarrow$      & Deg.$\downarrow$  & PSNR$\uparrow$  & SSIM$\uparrow$  \\ \hline
			Bicubic         & 0.4834 &148.87  &10.767 & 49.60 & 25.377 & 0.6985  \\
			RCAN*~\cite{zhang2018rcan}   &    0.4159    &   93.66      &    9.907     &     \blueud{38.45}    &   \redbf{27.24}     &   \redbf{0.7533}    \\
			ESRGAN*~\cite{wang2018esrgan}     & \blueud{0.4127} & \blueud{49.20} & \blueud{4.099} & \blueud{51.21}  & 23.74   &0.6319  \\
			Super-FAN~\cite{bulat2018super}  & 0.4791 &139.49 & 10.828  & 49.14   & 25.28   & \blueud{0.7033} \\ \hline
			\textbf{GFP-GAN (ours)} &  \redbf{0.3653}     &    \redbf{42.36}    &   \redbf{4.078}     &    \redbf{34.67}    &   25.04     &    0.6744    \\ \hline
			GT                 & 0    & 43.43  & 4.292     & 0  & $\infty$    & 1    \\ \hline
	\end{tabular}}
	\vspace{-0.5cm}
\end{table}

Qualitative results are presented in Fig.~\ref{fig:celeb_compare} and Fig.~\ref{fig:celeb_sr_compare}. 1) Thanks to the powerful generative facial prior, our GFP-GAN recovers faithful details in the eyes (pupils and eyelashes), teeth, \etc. 2) Our method treats faces as whole in restoration and could also generate realistic hair, while previous methods that rely on component dictionaries (DFDNet) or parsing maps (PSFRGAN) fail to produce faithful hair textures (2nd row, Fig.~\ref{fig:celeb_compare}). 3) GFP-GAN is capable of retaining fidelity, \eg, it produces natural closed mouth  without forced addition of teeth as PSFRGAN does (2nd row, Fig.~\ref{fig:celeb_compare}). And in Fig.~\ref{fig:celeb_sr_compare}, GFP-GAN also restores reasonable eye gaze direction.

\begin{table}[]
	\small
	\centering
	\caption{Quantitative comparison on the \textit{real-world} \textbf{LFW}, \textbf{CelebChild}, \textbf{WebPhoto}. \redbf{Red} and \blueud{blue} indicates the best and the second best performance.  `*' denotes finetuning on our training set. Deg. represents the identity distance.}
	\vspace{-0.3cm}
	\label{tab:real_test}
	\tabcolsep=0.1cm
	\scalebox{0.95}{
		\hspace{-0.5cm}
		\begin{tabular}{c|cc|cc|cc}
			\hline
			Dataset       & \multicolumn{2}{c|}{\textbf{LFW-Test}} & \multicolumn{2}{c|}{\textbf{CelebChild}}   & \multicolumn{2}{c}{\textbf{WebPhoto}} \\ 
			Methods       & FID$\downarrow$  & NIQE $\downarrow$  & FID$\downarrow$  & NIQE $\downarrow$   &FID$\downarrow$  & NIQE $\downarrow$   \\ \hline
			Input         & 137.56 &11.214 & 144.42 & 9.170 & 170.11  & 12.755  \\
			DeblurGANv2*~\cite{kupyn2019deblurganv2}    &   57.28    &  4.309      &   110.51     &   4.453     &    100.58    &    \blueud{4.666}  \\
			Wan \etal.~\cite{wan2020bringing}  &   73.19    &   5.034     &   115.70     &   4.849     &    100.40     &     5.705   \\
			HiFaceGAN~\cite{yang2020hifacegan}     &   64.50    &   4.510     &   113.00     &   4.855     &    116.12    &     4.885  \\
			DFDNet~\cite{li2020dfdnet}        &   62.57    &   \blueud{4.026}     &   111.55     &   \blueud{4.414}     &    100.68     &     5.293   \\
			PSFRGAN~\cite{chen2020psfrgan}      &   \blueud{51.89}    &   5.096     &   \blueud{107.40}     &   4.804     &    88.45     &     5.582  \\ \hline
			mGANprior~\cite{gu2020mGANprior}&   73.00    &   6.051     &   126.54     &   6.841      &    120.75     &     7.226   \\
			PULSE~\cite{menon2020pulse}&   64.86     &   5.097      &   \redbf{102.74}      &   5.225      &   \redbf{86.45}     &     5.146  \\\hline
			\textbf{GFP-GAN (ours)} &   \redbf{49.96}     &   \redbf{3.882}    &   111.78     &     \redbf{4.349}   &   \blueud{87.35}     &    \redbf{4.144}    \\ \hline
	\end{tabular}}
	\vspace{-0.5cm}
\end{table}

\noindent\textbf{Real-World LFW, CelebChild and WedPhoto-Test.}
To test the generalization ability, we evaluate our model on three different real-world datasets.
The quantitative results are show in Table.~\ref{tab:real_test}. 
Our GFP-GAN achieves superior performance on all the three real-world datasets, showing its remarkable generalization capability.  Although PULSE~\cite{menon2020pulse} could also obtain high perceptual quality (lower FID scores), it could not retain the face identity as shown in Fig~\ref{fig:real_comparison}.


The qualitative comparisons are shown in Fig.~\ref{fig:real_comparison}.
GFP-GAN could jointly conduct face restoration and color enhancement for real-life photos with the powerful generative prior. 
Our method could produce plausible and realistic faces on complicated real-world degradation while other methods fail to recover faithful facial details or produces artifacts (especially in WebPhoto-Test in Fig~\ref{fig:real_comparison}).
Besides the common facial components like eyes and teeth, GFP-GAN also perform better in hair and ears, as the GFP prior takes the whole face into consideration rather than separate parts.
With SC-SFT layers, our model is capable of achieving high fidelity. As shown in the last row of Fig.~\ref{fig:real_comparison}, most previous methods fail to recover the closed eyes, while ours could successfully restore them with fewer artifacts.
%


\subsection{Ablation Studies}\label{subsec:image_to_image_translation}

\begin{figure}[!t]
	\small
	\vspace{-0.4cm}
	\begin{center}
		\includegraphics[width=\linewidth]{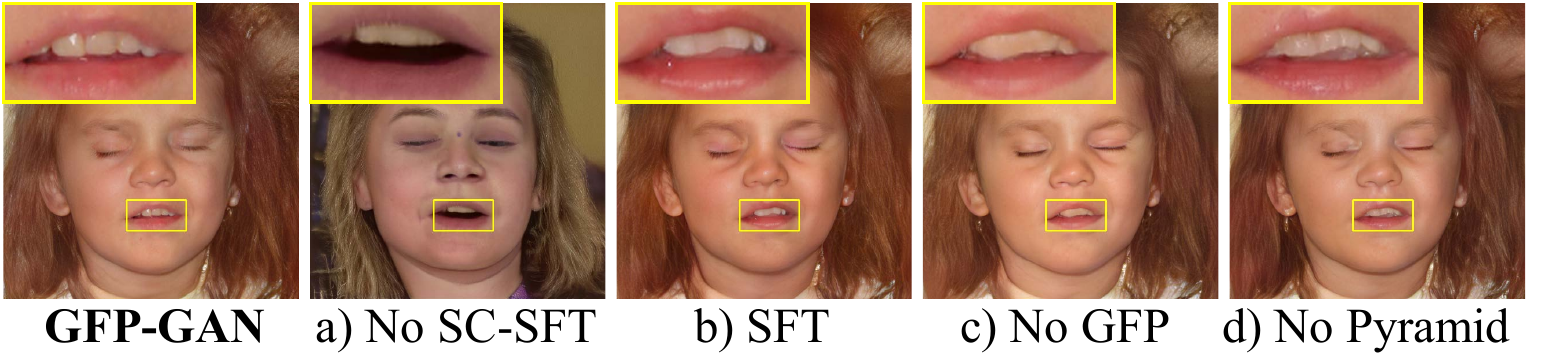}
	\end{center}
	\vspace{-0.7cm}
	\caption{Ablation studies on CS-SFT layers, GFP prior and pyramid restoration loss. \textbf{Zoom in for best view.}}
	\label{fig:ablation_a}
	\vspace{-0.2cm}
\end{figure}

\noindent\textbf{CS-SFT layers.}
As shown in Table.~\ref{tab:ablation} [configuration a)] and Fig.~\ref{fig:ablation_a}, when we remove spatial modulation layers, \ie, only keep the latent code mapping without spatial information, the restored faces could not retain face identity even with identity-preserving loss (high LIPS score and large Deg.). Thus, the multi-resolution spatial features used in CS-SFT layers is vital to preserve fidelity. 
When we switch CS-SFT layers to simple SFT layers [configuration b) in Table.~\ref{tab:ablation}], we observe that 1) the perceptual quality degrades on all metrics and 2) it preserves stronger identity (smaller Deg.), as the input image features impose influence on all the modulated features and the outputs bias to the degraded inputs, thus leading to lower perceptual quality. By contrast, CS-SFT layers provide a good balance of realness and fidelity by modulating a split of features. 
 
\noindent\textbf{Pretrained GAN as GFP.}
Pretrained GAN provides rich and diverse features for restoration. A performance drop is observed if we do not use the generative facial prior, as shown in Table.~\ref{tab:ablation}  [configuration c)] and Fig.~\ref{fig:ablation_a}. 
 
\noindent\textbf{Pyramid Restoration Loss.}
Pyramid restoration loss is employed in the degradation removal module and strengthens the restoration ability for complicated degradation in the real world. Without this intermediate supervision, the multi-resolution spatial features for subsequent modulations may still have degradation, resulting in inferior performance, as shown in Table.~\ref{tab:ablation} [configuration d)]  and Fig.~\ref{fig:ablation_a}.

\begin{figure}[!t]
	\small
	\begin{center}
		\includegraphics[width=\linewidth]{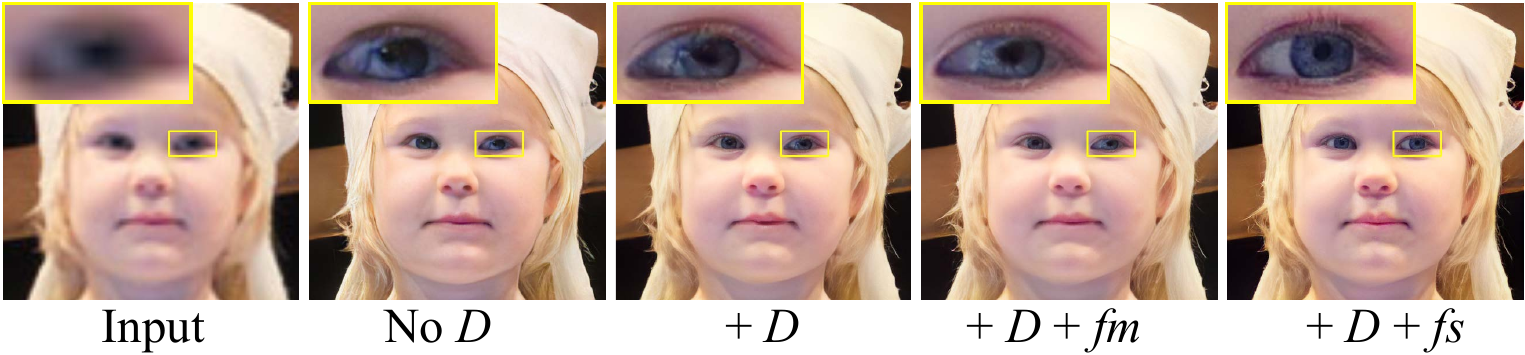}
	\end{center}
	\vspace{-0.7cm}
	\caption{Ablation studies on facial component loss. In the figure, \textit{D}, \textit{fm}, \textit{fs} denotes component discriminator, feature matching loss and feature style loss, respectively.}
	\label{fig:ablation_b}
	\vspace{-0.2cm}
\end{figure}

\noindent\textbf{Facial Component Loss.}
We compare the results of 1) removing all the facial component loss, 2) only keeping the component discriminators, 3) adding extra feature matching loss as in ~\cite{wang2017pix2pixHD}, and 4) adopting extra feature style loss based on Gram statistics~\cite{gatys2016style}. It is shown in Fig~\ref{fig:ablation_b} that component discriminators with feature style loss could better capture the eye distribution and restore the plausible details.

\begin{table}[!t]
	\vspace{-0.2cm}
	\small
	\centering
	\caption{Ablation study results on \textbf{CelebA-Test} under blind face restoration.}
	\vspace{-0.2cm}
	\label{tab:ablation}
	\tabcolsep=0.1cm
	\scalebox{0.95}{
		\hspace{-0.5cm}
		\begin{tabular}{l|ccc|c}
			\hline
			Configuration       & LPIPS$\downarrow$ & FID$\downarrow$  &NIQE $\downarrow$      & Deg.$\downarrow$   \\ \hline
			Our GFP-GAN with SC-SFT  & \textbf{0.3646}  & \textbf{42.62}  & \textbf{4.077} & \textbf{34.60} \\ \hline
			a) No spatial modulation    &    0.550 ($\uparrow$)   &  60.44  ($\uparrow$)    &  4.183 ($\uparrow$)    &    74.76 ($\uparrow$)    \\
			b) Use SFT   & 0.387 ($\uparrow$)  &  47.65 ($\uparrow$)   & 4.146($\uparrow$)  & 34.38 ($\downarrow$)    \\ \hline
			c) w/o GFP  &  0.379 ($\uparrow$)    &   48.47 ($\uparrow$)     &  4.153  ($\uparrow$) &   35.04  ($\uparrow$)    \\ \hline
			d) $-$ Pyramid Restoration Loss & 0.369 ($\uparrow$)    & 45.17 ($\uparrow$)& 4.284  ($\uparrow$)&35.50  ($\uparrow$)    \\ \hline
	\end{tabular}}
	\vspace{-0.1cm}
\end{table}

\begin{figure}[]
	\vspace{-0.6cm}
	\begin{center}
		\includegraphics[width=\linewidth]{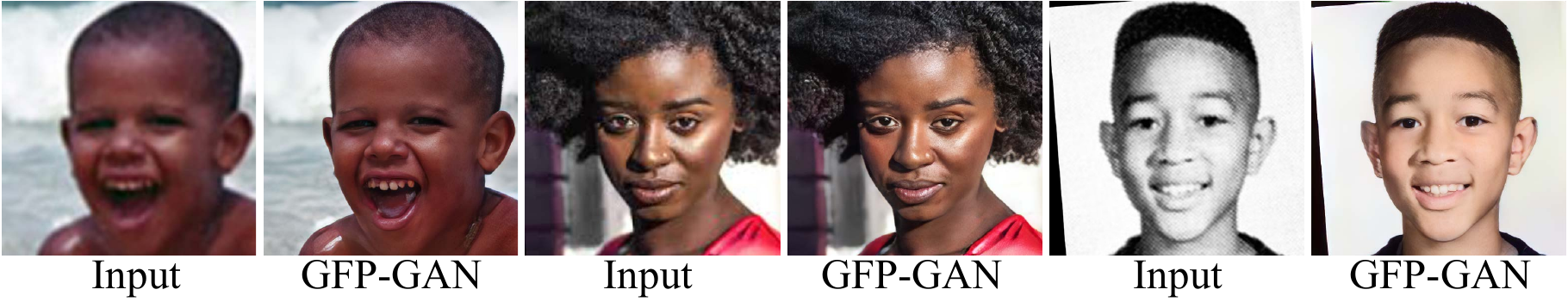}
	\end{center}
	\vspace{-0.8cm}
	\caption{Results on dark-skinned faces.}
	\label{fig:dark-skinned}
	\vspace{-0.4cm}
\end{figure}

\begin{figure}[!t]
	\small
	\begin{center}
		\includegraphics[width=\linewidth]{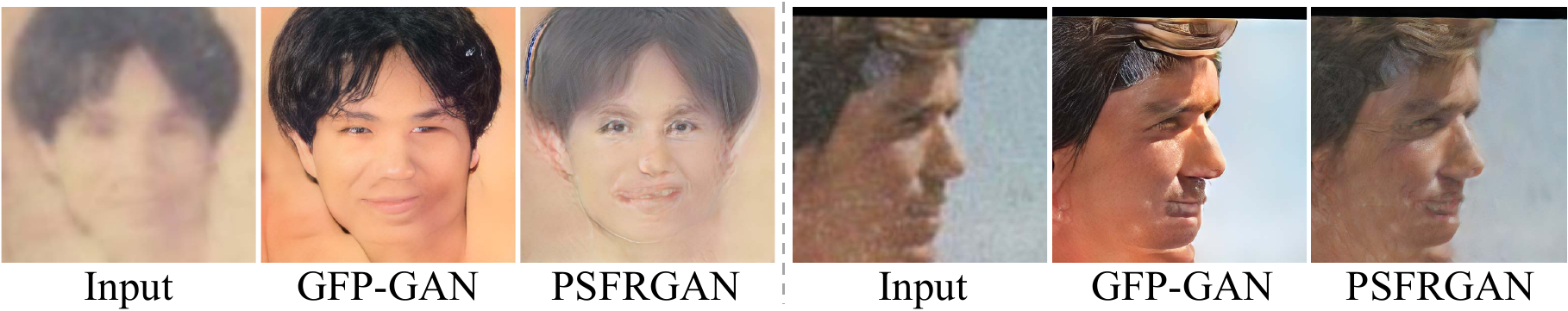}
	\end{center}
	\vspace{-0.7cm}
	\caption{Limitations of our model. The results of PSFRGAN~\cite{chen2020psfrgan} are also presented.}
	\label{fig:limitations}
	\vspace{-0.5cm}
\end{figure}

\vspace{-0.3cm}
\subsection{Discussion and Limitations}\label{subsec:limitations}
\noindent{\textbf{Training bias.}}
Our method performs well on most dark-skinned faces and various population groups (Fig.~\ref{fig:dark-skinned}), as our method uses both the pretrained GAN and input image features for modulation. Beside, we employ reconstruction loss and identity preserving loss to restrict the outputs to retain fidelity with inputs.
However, when input images are gray-scale, the face color may have a bias (last example in Fig.~\ref{fig:dark-skinned}), as the inputs do not contain sufficient color information. Thus, a diverse \textit{and} balanced dataset is in need. 

\noindent\textbf{Limitations.}
%
As shown in Fig.~\ref{fig:limitations}, when the degradation of real images is severe, the restored facial details by GFP-GAN are twisted with artifacts. Our method also produces 
unnatural results for very large poses. 
This is because the synthetic degradation and training data distribution are different from those in real-world. 
One possible way is to learn those distributions from real data instead of merely using synthetic data, which is left as future work.

%

\vspace{-0.2cm}
\section{Conclusion}
\vspace{-0.2cm}
We have proposed the GFP-GAN framework that leverages the rich and diverse generative facial prior for the challenging blind face restoration task. 
This prior is incorporated into the restoration process with channel-split spatial feature transform layers, allowing us to achieve a good balance of realness and fidelity. 
%
Extensive comparisons demonstrate the superior capability of GFP-GAN in joint face restoration and color enhancement for real-world images, outperforming prior art.

\newpage
{\small
\bibliographystyle{ieee_fullname}
\bibliography{bib}
}

\end{document}